\documentclass[letterpaper, 10 pt, journal, twoside]{IEEEtran}

\usepackage{xcolor}
\usepackage{graphicx}
\usepackage{multirow}
\usepackage{boldline}
\usepackage{pifont}
\usepackage{url}
\usepackage[symbol]{footmisc}
\usepackage{amsmath}
\usepackage{amssymb}
\newcommand{\cmark}{\ding{51}}
\newcommand{\xmark}{\ding{55}}

\begin{document}
\title{GSFusion: Online RGB-D Mapping Where Gaussian Splatting Meets TSDF Fusion}

\author{Jiaxin Wei$^{1,3}$ and Stefan Leutenegger$^{1,2,3}$
\thanks{$^{1}$Smart Robotics Lab, Technical University of Munich
{\tt\small \{jiaxin.wei,stefan.leutenegger\}@tum.de}}%
\thanks{$^{2}$Smart Robotics Lab, Imperial College London}%
\thanks{$^{3}$Munich Institute of Robotics and Machine Intelligence (MIRMI)}%
}


\maketitle

\begin{abstract}
Traditional volumetric fusion algorithms preserve the spatial structure of 3D scenes, which is beneficial for many tasks in computer vision and robotics. However, they often lack realism in terms of visualization. Emerging 3D Gaussian splatting bridges this gap, but existing Gaussian-based reconstruction methods often suffer from artifacts and inconsistencies with the underlying 3D structure, and struggle with real-time optimization, unable to provide users with immediate feedback in high quality. One of the bottlenecks arises from the massive amount of Gaussian parameters that need to be updated during optimization. Instead of using 3D Gaussian as a standalone map representation, we incorporate it into a volumetric mapping system to take advantage of geometric information and propose to use a quadtree data structure on images to drastically reduce the number of splats initialized. In this way, we simultaneously generate a compact 3D Gaussian map with fewer artifacts and a volumetric map on the fly. Our method, GSFusion, significantly enhances computational efficiency without sacrificing rendering quality, as demonstrated on both synthetic and real datasets. Code will be available at \url{https://github.com/goldoak/GSFusion}.
\end{abstract}


%
\IEEEpeerreviewmaketitle

\section{Introduction}
\label{sec:intro}

\begin{figure}[]
  \centering
5   \includegraphics[width=\linewidth]{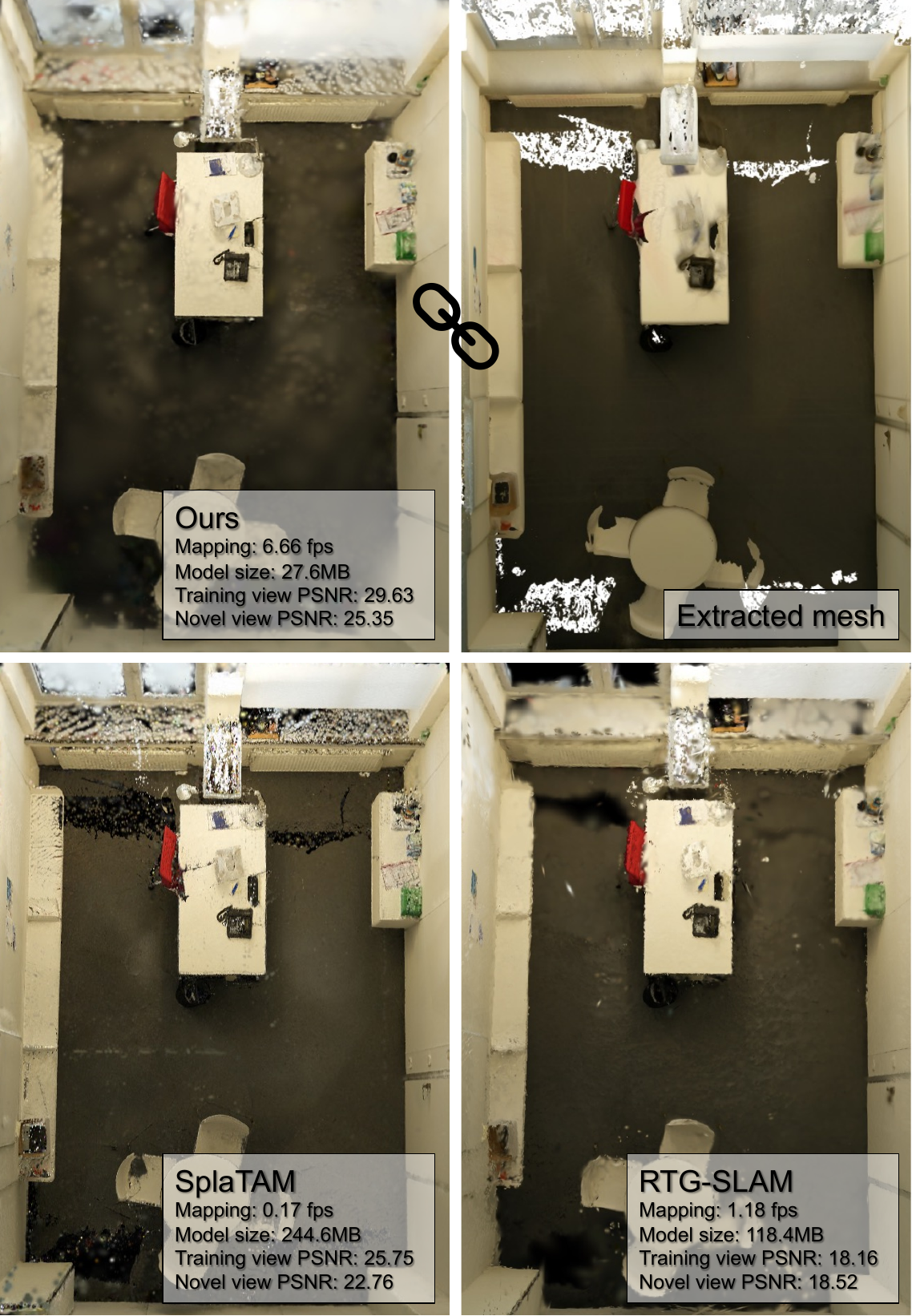}
   \caption{Comparison of methods on a real scene (8b5caf3398) from ScanNet++ dataset~\cite{scannetpp}. All the reported results are obtained from a single Nvidia RTX 3060 GPU.}
   \label{fig:front_fig}
\end{figure}

\IEEEPARstart{O}{nline} volumetric mapping~\cite{kinectfusion,niessner2013real,voxblox,se2,steinbrucker2013large} integrates depth data from sensors into a volumetric grid over time, where each voxel typically stores either an occupancy value or a Truncated Signed Distance Field (TSDF), i.e., truncated signed distance to the nearest surface. It creates an accurate geometric map for understanding the 3D structure of the environment, useful for robot navigation, path planning, and spatial reasoning. Color data can be fused into the same volumetric grid to add textures to the map, enabling applications that require both spatial and visual information, such as VR/AR experiences. However, naive color fusion usually results in low photo-realism and incomplete maps due to occlusions, holes in depth maps, and neglect of material properties, such as reflection and transparency.

To enhance the visual fidelity of the 3D map, some previous works utilize Neural Radiance Fields (NeRFs)~\cite{nerf}, a neural implicit representation, to model the scene via a simple multilayer perceptron (MLP)~\cite{imap} or a multi-resolution feature grid~\cite{nice-slam}. The visual improvement is largely attributed to the differentiable volume rendering that incorporates transparency and directional dependency, but it is also computationally expensive, making NeRF-based reconstruction methods challenging to achieve real-time performance.

Recently, a new promising technique for radiance field rendering, known as 3D Gaussian Splatting (3DGS)~\cite{3dgs}, has emerged. In contrast to NeRF, 3DGS employs an explicit 3D Gaussian representation along with a tile-based rasterizer, showcasing state-of-the-art visual quality while maintaining high rendering speed. It also speeds up the reconstruction process, allowing it to compete with the fastest NeRF-based methods, but the original paradigm is designed for offline operation, which still takes approximately 10-30 minutes to batch-optimize over a small-scale dataset ($\sim$300 images). 

Nonetheless, with comparable visual quality and fast rendering speed, several concurrent works~\cite{monogs, gs-slam, gaussian-slam, splatam, rtg-slam} build RGB-D SLAM systems upon this novel representation and propose different strategies attempting to fulfill online optimization demands. Instead of relying on the sparse point cloud generated by the Structure-from-Motion (SfM)~\cite{sfm} process, an incremental Gaussian initialization is employed to better handle sequential RGB-D inputs. This modification helps reduce some floating artifacts, but those methods usually rely on dense pixel-wise sampling when initializing a new Gaussian in 3D space, leading to a substantial storage overhead and hindering the overall efficiency. As a result, current Gaussian-based SLAM systems still struggle to reach true real-time. RTG-SLAM~\cite{rtg-slam} explicitly imposes several restrictions on 3D Gaussians to reduce the number of parameters to be optimized and the number of pixels to be rendered, enabling real-time performance. However, it is not robust against missing depths, thereby leaving holes in the resulting map, particularly around windows and mirrors.

In this paper, we argue that limiting the number of 3D Gaussians added to the scene is crucial for maintaining a balance between visual quality and efficiency during scanning. Moreover, instead of using 3D Gaussians as the only map representation, we combine 3DGS with a traditional volumetric fusion approach. The benefits are twofold. First, we can fully exploit the spatial structure of volumetric grids to avoid duplicating 3D Gaussians within a neighborhood. Second, volumetric maps are more established compared to 3DGS, making them suitable for a wider range of downstream robotics tasks. Specifically, we develop our GSFusion on top of Supereight2~\cite{se2}, a high-performance volumetric mapping system, to 
simultaneously generate a 3D Gaussian map and a TSDF map on the fly. The key to the efficiency of our method lies in a quadtree decomposition of input images, which helps determine where to add a new Gaussian. Each input RGB image is subdivided into cells of varying sizes based on contrast. A 3D Gaussian is then initialized at the center of a cell after checking its vicinity using geometric information. Our proposed method not only significantly reduces the total number of Gaussian primitives, resulting in a clean and compact 3D Gaussian map, but also eases the online optimization process with much less computation needed (see Fig.~\ref{fig:front_fig}). Extensive experiments on both synthetic and real datasets demonstrate the effectiveness of our GSFusion. Additionally, we conduct detailed ablation studies on several design choices to offer insights on how to maximize its advantages for different purposes. Our main contributions are summarized as follows:
\begin{itemize}
    \item We present GSFusion, a hybrid mapping system that combines Gaussian splatting and TSDF fusion to generate two types of maps simultaneously in real time.
    \item We utilize a quadtree division scheme, as well as the volumetric grid structure, to explicitly control the number of new Gaussians added to the scene, thereby optimizing both model size and computational efficiency.
    \item We conduct comprehensive experiments on two benchmark datasets, showing that our method effectively balances mapping frequency and rendering quality compared to previous Gaussian-based approaches.
\end{itemize}

\begin{figure*}[]
  \centering
   \includegraphics[width=\linewidth]{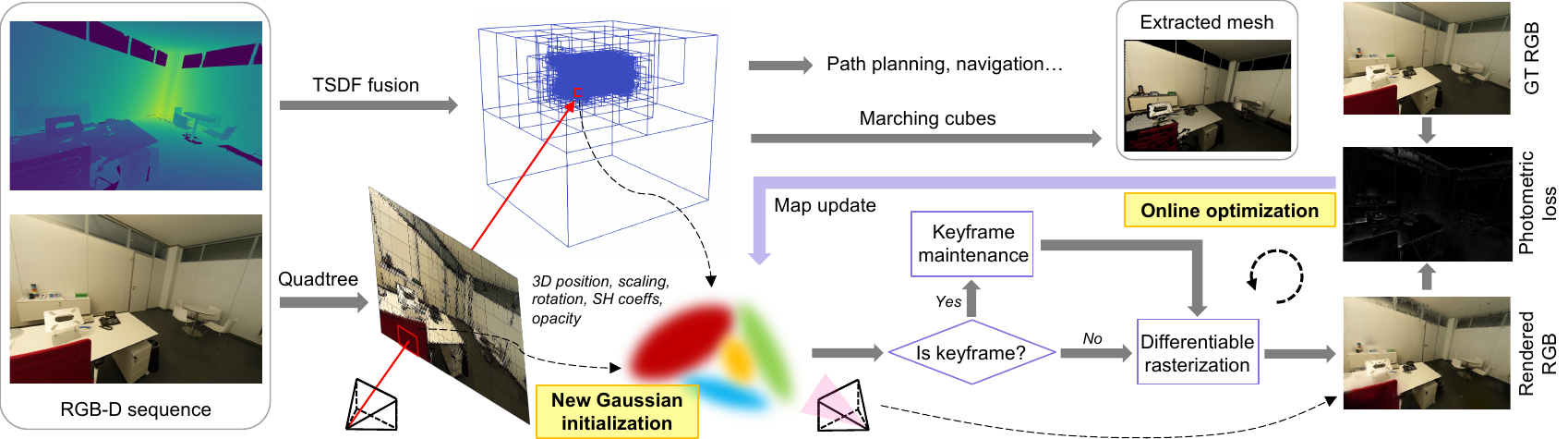}
   \caption{System overview of our proposed GSFusion. At each time step, it takes a pair of RGB-D images as input. The depth data is fused into an octree-based TSDF grid to capture geometric structure while the RGB image is segmented using quadtree based on contrast. A new 3D Gaussian is then initialized at the back-projected center of a quadrant if there are no adjacent Gaussians by checking its nearest voxel. We optimize Gaussian parameters on the fly by minimizing the photometric loss between the rendered image and input RGB. Additionally, we maintain a keyframe set to deal with the forgetting problem. After scanning, the system provides both a volumetric map and a 3D Gaussian map for subsequent tasks.}
   \label{fig:pipeline}
\end{figure*}

\section{Related Work}
\label{sec:related_work}

\subsection{Traditional Volumetric Mapping}
Extensive research has been conducted in the field of RGB-D reconstruction. A seminal work by Curless and Levoy~\cite{tsdf} introduces the use of truncated signed distance fields for volumetric mapping, which has since become one of the popular map representations for achieving high-quality reconstruction due to its ability to effectively model continuous surfaces. Building upon this representation, KinectFusion~\cite{kinectfusion} proposes efficient implementation and optimized volume integration, making real-time 3D reconstruction feasible. To tackle scalability challenges in large environments, voxel hashing techniques, as demonstrated by Niessner et al.~\cite{niessner2013real} and in Voxblox\cite{voxblox}, are used to dynamically allocate and access 3D voxels as needed. Researchers also explore the integration of hierarchical data structures, such as octrees utilized in Supereight2~\cite{se2} and in the work by Steinbrucker et al.~\cite{steinbrucker2013large}. This approach allows for adaptive resolution and more efficient memory usage, enabling volumetric reconstruction to be applied to even larger and more complex scenes. Though these methods are good at reconstructing geometrically accurate surfaces beneficial for downstream robotics tasks, they lack photo-realism in maps, which is desirable for specific inspection missions. Our GSFusion builds on the efficient TSDF fusion of Supereight2 and enhances it by integrating Gaussian splatting, resulting in a high-quality map suitable for visualization.

\subsection{NeRF-based Reconstruction}
Recent advancements in neural radiance fields have inspired several NeRF-based RGB-D SLAM systems. iMAP~\cite{imap} pioneers this approach by using a single MLP to model the scene, while NICE-SLAM~\cite{nice-slam} enhances scalability through hierarchical feature grids. Other notable methods, including Vox-Fusion~\cite{vox-fusion}, ESLAM~\cite{eslam}, Co-SLAM~\cite{co-slam} and Point-SLAM~\cite{point-slam} explore different data structures towards efficient mapping. These methods achieve impressive results in terms of visual quality, however, they struggle with real-time performance and high memory costs due to the computational demands of volume rendering.

\subsection{Gaussian-based Reconstruction}
Gaussian splatting~\cite{3dgs} enables high-quality image rendering through differentiable rasterization, offering a significant speed advantage over NeRF. Therefore, several concurrent works employ 3D Gaussians as scene representations in dense RGB-D SLAM systems. GS-SLAM~\cite{gs-slam} introduces an adaptive 3D Gaussian expansion strategy that uses opacity to dynamically add or remove Gaussians. Gaussian-SLAM~\cite{gaussian-slam} splits the scene into sub-maps and seeds two Gaussians at the same time for each sampled point so that no cloning or pruning is needed as in \cite{3dgs}. SplaTAM~\cite{splatam} ensures isotropic Gaussians and adds new Gaussians according to a densification mask derived from rendered depth and silhouette images. MonoGS~\cite{monogs} places Gaussians around depth measurements with different levels of variance and prunes unnecessary Gaussians through geometric verification. RTG-SLAM\cite{rtg-slam} forces each Gaussian to be either opaque or transparent and explicitly adds Gaussians for three types of pixels (i.e., newly observed or with large color/depth errors). To enhance efficiency, RTG-SLAM only optimizes and renders unstable Gaussians, but it struggles with missing depth data, leading to incomplete reconstructions in challenging scenarios like windows and mirrors. Despite these advances, real-time, photo-realistic reconstruction with low memory cost remains a significant challenge, which our proposed method aims to address through a more compact Gaussian representation and highly efficient optimization strategy.

\section{Method}
\label{sec:method}

Our mapping system is illustrated in Fig.~\ref{fig:pipeline}. We begin with a brief review of the two scene representations involved in our method in Section~\ref{sec:primitive}. Next, we describe in detail the process of incremental mapping based on these primitives in Sec.~\ref{sec:mapping}, including TSDF fusion, new Gaussian initialization, and online optimization of Gaussian parameters.

\subsection{Hybrid Scene Representation}
\label{sec:primitive}
\subsubsection{Octree-based TSDF Grid}
An octree is a spatial data structure, commonly used to partition 3D space into eight octants for the sake of computational efficiency. In Supereight2, each leaf node contains a contiguous block of $8^3$ voxels. This converts the volumetric map into a collection of unordered, sparsely distributed voxel blocks, with the octree serving as a spatial index to retrieve data based on integer coordinates. Here efficient tree traversal is achieved using Morton coding, where interleaved bits from the 3D coordinates form a unique identifier for each node. This coding scheme implicitly defines the tree structure as it recursively indicates the positions of parent nodes in coarser levels. Note that this mapping system supports flexible data types, such as TSDF and occupancy. We choose single-resolution TSDF for simplicity of implementation. Hence, each voxel stores a truncated signed distance to the closest surface and negative values represent voxels behind the surface.

\subsubsection{3D Gaussian}
To achieve high-quality rendering, 3D Gaussians are also chosen as the scene primitive because they are differentiable and can be easily projected into 2D space enabling fast $\alpha$-blending. A 3D Gaussian can be viewed as an ellipsoid, where the $3\times 3$ covariance matrix $\mathbf{\Sigma}$ defines the shape and orientation of the ellipsoid in 3D space. Let us assume that a 3D Gaussian is centered at $\mathbf{p}_g$, then its definition is given as follows:
\begin{gather}
    G(\mathbf{x}) = \exp\left({-\frac{1}{2}(\mathbf{x} - \mathbf{p}_g)^T\mathbf{\Sigma}^{-1}(\mathbf{x} - \mathbf{p}_g)}\right),~\mathbf{x} \in \mathbb{R}^3, \\
    \mathbf{\Sigma} = \mathbf{RS}\mathbf{S}^T\mathbf{R}^T,
\end{gather}
where $\mathbf{R} \in \mathbb{R}^{3\times 3}$ and $\mathbf{S} \in \mathbb{R}^{3\times 3}$ denotes the rotation and scaling of the 3D Gaussian, respectively. Given the rigid transformation $\mathbf{T}_{WC_k} \in \text{SE}(3)$ from the camera coordinate frame to the world coordinate frame at time step $k$, the projected 2D splat is defined as:
\begin{gather}
    \hat{G}(\mathbf{u}) = \exp\left({-\frac{1}{2}(\mathbf{u} - \boldsymbol\mu)^T\hat{\mathbf{\Sigma}}^{-1}(\mathbf{u} - \boldsymbol\mu)}\right),~\mathbf{u} \in \mathbb{R}^2, \label{Eq:2d_splat}\\
    \boldsymbol\mu = \pi\left(\mathbf{T}_{WC_k}^{-1}\mathbf{p}_g\right), \\
    \hat{\mathbf{\Sigma}} = \mathbf{J}\mathbf{W}\mathbf{\Sigma}\mathbf{W}^T\mathbf{J}^T,
\end{gather}
where $\pi(\cdot)$ is used to perform dehomogenization and perspective projection in order. $\mathbf{W}$ denotes the rotational part of $\mathbf{T}_{WC_k}^{-1}$ and $\mathbf{J}$ is the Jacobian of the affine approximation of the projective transformation that converts camera coordinates to ray coordinates. Apart from the 3D position $\mathbf{p}_g$, rotation $\mathbf{R}$ and scaling $\mathbf{S}$, each 3D Gaussian also contains an opacity value $\alpha \in [0, 1]$ for blending and spherical harmonics (SH) coefficients to capture the view-dependent color $\mathbf{c}$.

\subsection{Incremental Mapping}
\label{sec:mapping}

Our real-time incremental mapping process consists of three main stages. First, it performs TSDF fusion using the input depth frame to update necessary voxels. Then, we assign new Gaussians in 3D space by utilizing both visual and geometric information to avoid the explosion of primitives. Finally, we design an effective keyframe maintenance strategy to optimize Gaussian parameters on the fly, reaching a balance between rendering quality and computational efficiency.

\subsubsection{TSDF Fusion}
The TSDF fusion process involves integrating new sensor data into the volumetric map. Therefore, it is important to determine which voxel blocks should be updated given the current depth frame. Since TSDF fields only encode surface information within the truncation bandwidth $\pm\epsilon$, a ray is marched along the line of sight for each pixel, and voxel blocks are allocated only if they fall into the specified $\epsilon$ band around the depth measurement. Supereight2~\cite{se2} employs a breadth-first allocation from top to bottom, taking advantage of Morton coding for fast voxel block allocation with minimal overhead. It is worth noting that after an initial phase, fewer voxel blocks need to be allocated in subsequent frames, as new blocks will mostly appear at frame borders or in previously unseen regions. Once allocation is completed, the integration of measurements is carried out in the same manner as KinectFusion~\cite{kinectfusion}. By projecting the voxel positions $\mathbf{p}_v$ into image coordinates, a new SDF value is obtained by calculating the difference between the measured distance from depth images and the metric distance from the camera center to the voxel. The new value is further normalized to a TSDF and fused with the previous value using a simple running weighted average (assuming an upper limit $w_{\text{max}}$ for weights). Mathematically:
\begin{align}
    \text{sdf} &= \mathbf{D}_k[\pi\left(\mathbf{T}_{WC_k}^{-1}\mathbf{p}_v\right)] - z_v^c, \\
    \text{tsdf} &= \begin{cases}
    \min(1, \frac{\text{sdf}}{\epsilon}), & \text{if sdf}>0,\\
    \max(-1, \frac{\text{sdf}}{\epsilon}), & \text{otherwise}.
    \end{cases} \\
    w_k &= \min\left(w_{\text{max}}, w_{k-1} + 1\right), \label{Eq:voxel_weight} \\
    \text{tsdf}_k &= \frac{\text{tsdf}_{k-1}w_{k-1} + \text{tsdf}~w_k}{w_{k-1} + w_k},
\end{align}
where $\mathbf{D}_k$ denotes the input depth map at time step $k$ and $z_v^c$ is introduced to represent the $z$-axis coordinate of $\mathbf{p}_v$ in the camera frame. Lastly, both the updated $w_k$ and $\text{tsdf}_k$ are stored in the corresponding voxel.

\begin{figure}[]
  \centering
   \includegraphics[width=\linewidth]{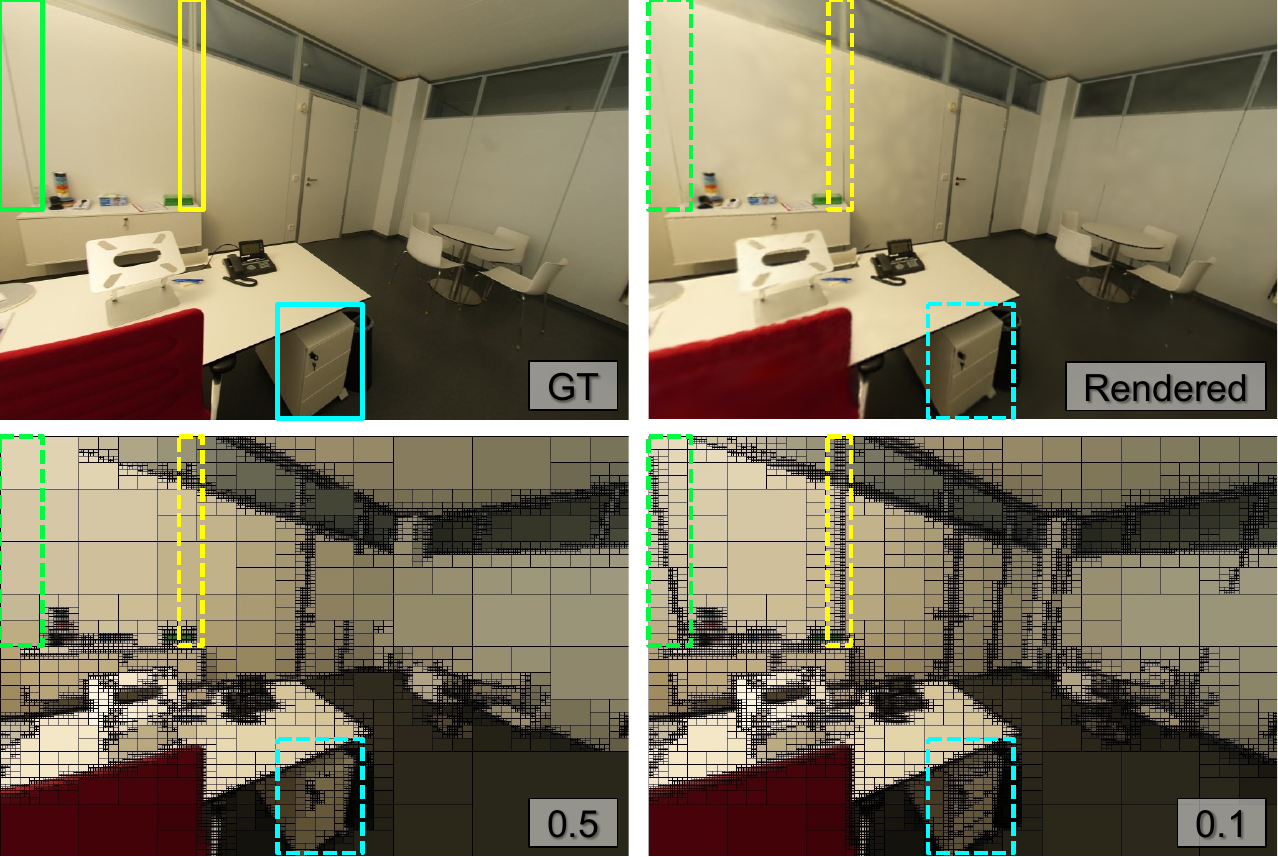}
   \caption{Effect of different quadtree thresholds. Top: input RGB image (left) and image rendered from the map created using 0.1 quadtree threshold (right). Bottom: RGB images segmented with different quadtree thresholds. Using stricter quadtree thresholds can help capture finer details, particularly thin edges caused by contrasts.}
   \label{fig:qtree}
\end{figure}

\subsubsection{New Gaussian Initialization}
In order to explicitly control the number of Gaussian primitives to be added to the map, we propose to use a quadtree scheme on each input RGB image. A quadtree is similar to an octree but operates in 2D space, with each node in the tree having either exactly four children or none at all (i.e., a leaf node). In our approach, we divide the image based on contrast so that each leaf node contains a specific subregion with its contrast lower than a pre-defined threshold. In Fig.~\ref{fig:qtree}, we demonstrate how different quadtree thresholds can impact partition results. Some intricate details, such as edges resulting from contrasts, can be easily overlooked during initialization while using a stricter threshold allows us to capture these details without the need for exhaustive pixel sampling. Assuming the center of a quadrant is $\mathbf{u}_q \in \mathbb{R}^2$, we can back-project it into the world coordinate frame using the depth map:
\begin{equation}
    \mathbf{p}_q = \mathbf{T}_{WC_k}\pi^{-1}(\mathbf{u}_q, \mathbf{D}_k[\mathbf{u}_q]),
\end{equation}
where $\pi^{-1}(\cdot)$ is the inverse of perspective projection. Before placing a new 3D Gaussian at $\mathbf{p}_q$, we first need to check its surroundings. This is to prevent duplicating multiple highly similar primitives within a neighborhood. Otherwise, the map will keep expanding even if no new areas are identified during the scanning process. Therefore, we simply query the nearest voxel to $\mathbf{p}_q$. If the voxel weight (see Eq.~\ref{Eq:voxel_weight}) is equal to 1, it indicates that it is a newly allocated voxel and we can safely initialize a new 3D Gaussian at $\mathbf{p}_q$ as follows:
\begin{equation*}
    \mathbf{R}_q = \mathbf{I},~~\mathbf{S}_q = \text{diag}\{d, d, d\},~~\alpha_q = 0.5,~~\mathbf{c}_q = \mathbf{I}_k[\mathbf{u}_q]
\end{equation*}
where $\mathbf{R}_q$ is a $3\times 3$ identity matrix and $\mathbf{S}_q$ is a $3\times 3$ diagonal matrix with each entry $d$ equalling to the back-projected length from the quadrant center to its corners. Note that the first three SH coefficients are initialized using the pixel color in image $\mathbf{I}_k$. Once the voxel weight exceeds 1, we then know that this location has already been visited before, and no new Gaussians should be assigned here. 

Our initialization strategy offers several significant benefits. It not only considerably reduces the map size, thus improving computational efficiency, but also maximizes the use of each primitive, resulting in fewer floating artifacts in the scene. Moreover, the compactness of our resulting map eliminates the need for further densification or pruning processes as required in the original 3DGS method~\cite{3dgs}, which is another computational bottleneck of online optimization.

\subsubsection{Online Optimization}
Now, we can directly optimize Gaussian parameters using the gradient descent method, thanks to differentiable rendering. We adopt a tile-based rasterizer which divides the screen into multiple tiles. 3D Gaussians lying in the current view frustum are sorted based on depth in each tile. As a result, a rendered RGB image $\hat{\mathbf{I}}_k$ can be obtained by blending $N$ sorted and splatted primitives (see Eq.~\ref{Eq:2d_splat}) that overlap at a pixel
\begin{equation}
    \hat{\mathbf{I}}_k[\mathbf{u}] = \sum_{i=1}^{N} \mathbf{c}_i \alpha_i \hat{G}_i(\mathbf{u}) \prod_{j=1}^{i-1} (1 - \alpha_j \hat{G}_j(\mathbf{u})).
\end{equation}
The objective function for optimization is then defined as
\begin{equation}
    L = \|\mathbf{I}_k - \hat{\mathbf{I}}_k\|_1, \label{Eq:loss}
\end{equation}
trying to minimize the photometric loss between the input RGB image and the rendered image. All the gradients for parameters are explicitly derived to reduce computational overhead.

At each time step, we execute the optimization in Eq.~\ref{Eq:loss} for several iterations. The more iterations we run, the better quality rendered image we can get from the current viewpoint. However, the online process cannot afford numerous iterations of first-order optimization. Additionally, it also leads to severe over-fitting issues, ultimately degrading the overall quality of the 3D Gaussian map. To balance the rendering quality and optimization time, we develop a simple yet effective online optimization strategy making use of a keyframe list. During our initialization stage, new Gaussians are assigned only to previously unobserved areas. Therefore, the number of new primitives initialized per frame naturally serves as an indicator of information gain from each input frame. Simply, we select a keyframe and add it to the list once the number of new primitives is greater than a certain threshold. Keyframes and non-keyframes will undergo $m$ and $n$ iterations of optimization, respectively. However, we spend less time on optimizing non-keyframes, reserving some resources to randomly selected keyframes from the maintained list (i.e., the first $(m-n)$ keyframes after shuffling). This random keyframe optimization helps mitigate forgetting issues without significantly increasing the computational burden. Similar to \cite{rtg-slam}, our system also supports offline optimization over the keyframe list after scanning.

\section{Experiments}
\label{sec:experiments}

We conduct comprehensive experiments on our mapping system using both synthetic and real datasets. The detailed experimental setup is first described in Sec.~\ref{sec:exp_setup}. We then analyze the mapping performance in Section~\ref{sec:exp_results} and validate our design choices in Section~\ref{sec:exp_ablation} through ablation studies.

\subsection{Experimental Setup}
\label{sec:exp_setup}
\subsubsection{Datasets}
We use ScanNet++~\cite{scannetpp} and Replica~\cite{replica} as our benchmark datasets. ScanNet++ is a real-world indoor dataset featuring high-quality RGB and depth images. More importantly, it provides two separate trajectories per scene for training and evaluation, allowing for comparisons of rendering quality on novel views. We evaluate four scenes from ScanNet++ (8b5caf3398, 39f36da05b, b20a261fdf, and f34d532901) as used in \cite{rtg-slam}. However, the sparse viewpoints along the training trajectories in ScanNet++ present a significant challenge for online optimization. In contrast, the Replica dataset, which consists of synthetic indoor scenes, offers more consistent trajectories and highly accurate RGB-D images, making it a good complement to ScanNet++. Considering the GPU memory, we downsample the high-resolution images ($1752\times 1168$) in ScanNet++ by a factor of 2 to prevent memory overflow.

\subsubsection{Baselines}
We compare our method with two state-of-the-art Gaussian SLAM approaches, SplaTAM~\cite{splatam} and RTG-SLAM~\cite{rtg-slam}, as they are currently the only open-source implementations supporting evaluation on the ScanNet++ dataset. We reproduce their results using the official codebases, strictly following their specified experimental parameters. For a fair comparison, we use ground truth camera poses across all methods.

\subsubsection{Evaluation Metrics}
To assess rendering quality, we consider three commonly used metrics: PSNR, SSIM, and LPIPS. Additionally, we evaluate computational efficiency and memory footprint by measuring the mapping frame rate (or FPS), GPU memory usage during mapping, and the model size of the resulting Gaussian map. All metrics are averaged over scenes in each dataset.

\subsubsection{Implementation Details}
All experiments in this paper are conducted on a desktop computer equipped with an Intel i7-13700 CPU and a single Nvidia RTX 3060 GPU. Our mapping system is implemented in C++ using the LibTorch framework, with differentiable rendering written in custom CUDA kernels for optimized performance.

For both datasets, we set the voxel size to 1cm, the quadtree threshold to 0.1, the keyframe threshold to 50, and the number of iterations for global optimization after scanning to 10. For the Replica dataset, we set $m=5$ and $n=3$ with an additional 2 iterations for random keyframe optimization. Since ScanNet++ has fewer frames per scene, we set $m=10$ and $n=1$ along with 9 iterations of random keyframe optimization. Due to the sparsity of viewpoints in ScanNet++, all frames are stored in the keyframe list.

\begin{table}[]
\caption{Rendering Performance on ScanNet++ Dataset}
\centering
\resizebox{\columnwidth}{!}{%
\begin{tabular}{c|ccccc}
\hlineB{2.5}
Global opt.                            & Method                    & Data splits   & PSNR$\uparrow$ & SSIM$\uparrow$ & LPIPS$\downarrow$ \\ \hlineB{2.5}
\multirow{6}{*}{\xmark} & \multirow{2}{*}{SplaTAM}  & Training view & \textbf{25.22} & \textbf{0.840} & \textbf{0.167}    \\
                                       &                           & Novel view    & \textbf{23.02} & 0.791          & \textbf{0.229}    \\ \cline{2-6} 
                                       & \multirow{2}{*}{RTG-SLAM} & Training view & 18.28          & 0.777          & 0.250             \\
                                       &                           & Novel view    & 18.69          & 0.750          & 0.303             \\ \cline{2-6} 
                                       & \multirow{2}{*}{Ours}     & Training view & 24.99          & 0.839          & 0.243             \\
                                       &                           & Novel view    & 22.76          & \textbf{0.794} & 0.313             \\ \hlineB{2.5}
\multirow{4}{*}{\cmark} & \multirow{2}{*}{RTG-SLAM} & Training view & 19.11          & 0.808          & 0.201             \\
                                       &                           & Novel view    & 19.61          & 0.778          & 0.259             \\ \cline{2-6} 
                                       & \multirow{2}{*}{Ours}     & Training view & \textbf{28.84} & \textbf{0.897} & \textbf{0.138}    \\
                                       &                           & Novel view    & \textbf{25.45} & \textbf{0.848} & \textbf{0.216}    \\ \hlineB{2.5}
\end{tabular}%
}
\label{tab:scannetpp_psnr}
\end{table}

\subsection{Mapping Performance}
\label{sec:exp_results}

We report the quantitative rendering results for ScanNet++ and Replica datasets in Table~\ref{tab:scannetpp_psnr} and Table~\ref{tab:replica_psnr}, respectively. Since both our method and RTG-SLAM support global optimization over the keyframe list after scanning, we directly compare these two methods using full settings, while no global optimization is used for a fair comparison with SplaTAM. Here RTG-SLAM applies 10 and 20 iterations of global optimization on ScanNet++ and Replica. Additionally, we compare the mapping efficiency and memory cost in Table~\ref{tab:exp_efficiency}.

\begin{figure*}[]
  \centering
   \includegraphics[width=0.85\linewidth]{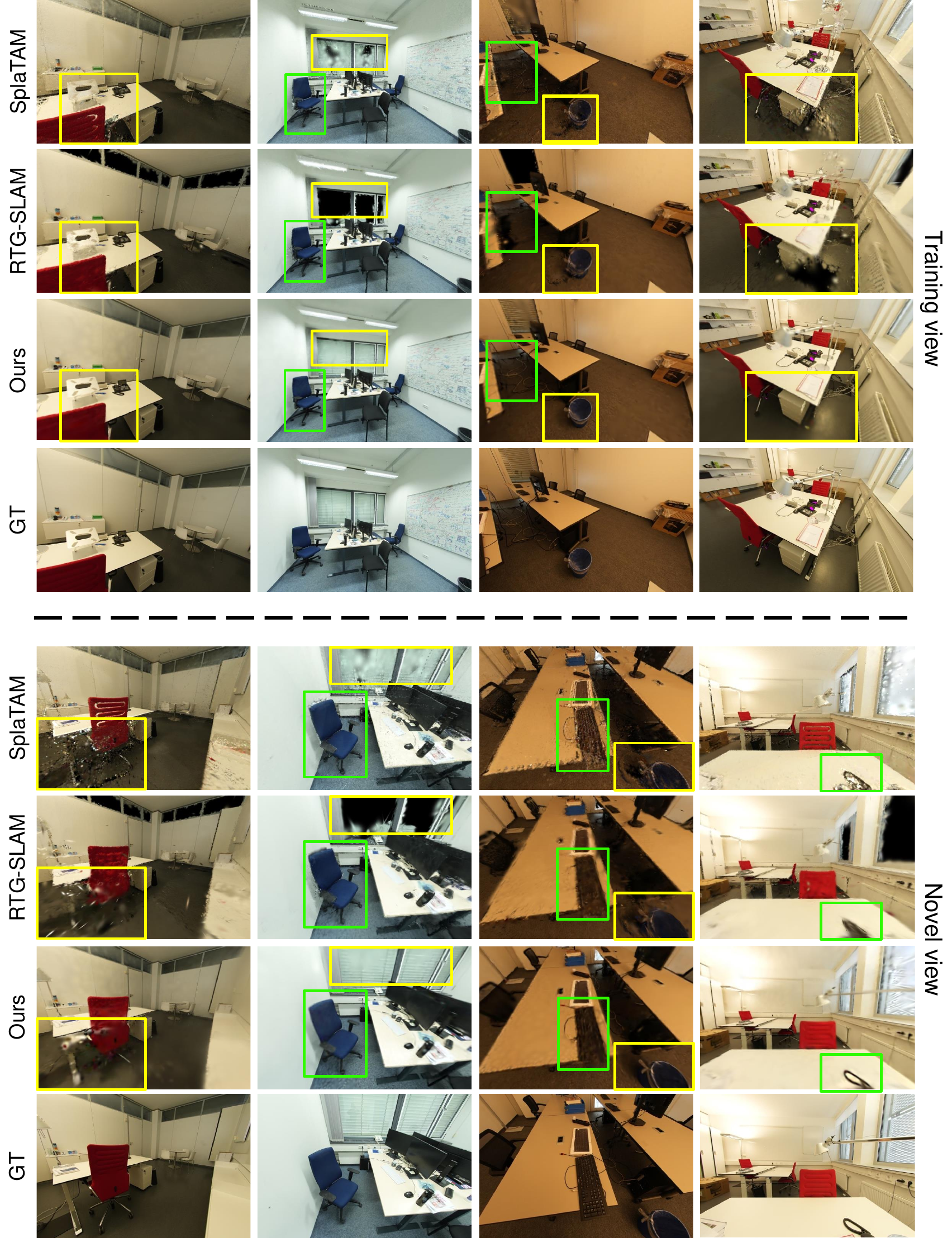}
   \caption{Qualitative rendering results from training and novel views on the ScanNet++ dataset. Zoom in for a clearer view.}
   \label{fig:scannetpp_vis}
\end{figure*}

Even though SplaTAM produces better rendering results on both datasets when there is no global optimization, it runs slowly ($<0.2$ fps) and consumes a large amount of GPU memory during mapping as it trades efficiency for quality. On the other hand, our method delivers comparable rendering results to RTG-SLAM on the Replica dataset and outperforms it on the ScanNet++ dataset. The performance discrepancy of RTG-SLAM lies in its inability to well handle the imperfections in real depth maps, especially in reflective or transparent areas with missing depth. This limitation is evident in Fig.~\ref{fig:scannetpp_vis}, where RTG-SLAM fails to model the windows in the scene. Meanwhile, our method can achieve even higher visual quality after only 10 iterations of global optimization, surpassing the other two methods on both synthetic and real datasets. Our GSFusion is also significantly more efficient in terms of mapping speed and memory usage. It runs at least 30x faster than SplaTAM and 5x faster than RTG-SLAM on ScanNet++ dataset with the most compact model size, which is at least 7x smaller than SplaTAM and 4x smaller than RTG-SLAM. The qualitative results on ScanNet++ are shown in Fig.~\ref{fig:scannetpp_vis}. Our GSFusion demonstrates fewer artifacts in reconstructed scenes from both training views and novel views. We suggest zooming in for a clearer view of details.

\begin{figure*}[]
  \centering
   \includegraphics[width=0.85\linewidth]{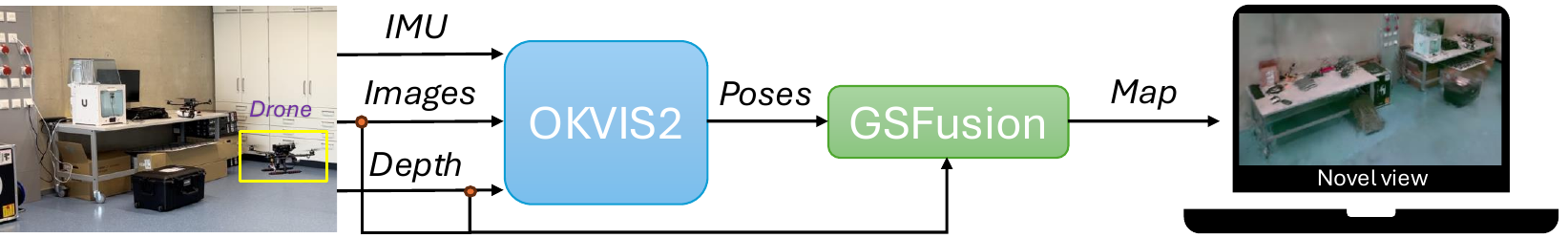}
   \caption{Workflow of our GSFusion applied to self-collected real-world drone data.}
   \label{fig:drone_fig}
\end{figure*}

We also demonstrate the effectiveness of our GSFusion system on real-world data captured by a drone equipped with an Intel RealSense D455 camera, with camera poses estimated using OKVIS2~\cite{okvis2}. The overall workflow is depicted in Fig.~\ref{fig:drone_fig}. Please refer to the supplementary video for more results.

\begin{table}[]
\caption{Rendering Performance on Replica Dataset}
\centering
\resizebox{0.8\columnwidth}{!}{%
\begin{tabular}{c|cccc}
\hlineB{2.5}
Global opt.                            & Method            & PSNR$\uparrow$ & SSIM$\uparrow$ & LPIPS$\downarrow$ \\ \hlineB{2.5}
\multirow{3}{*}{\xmark} & SplaTAM\footnotemark[2] & (32.56)        & (0.930)        & (0.064)           \\
                                       & RTG-SLAM          & \textbf{30.03} & 0.892          & 0.136             \\
                                       & Ours              & 28.64          & \textbf{0.900} & \textbf{0.103}    \\ \hlineB{2.5}
\multirow{2}{*}{\cmark} & RTG-SLAM          & 33.38          & 0.929          & 0.069             \\
                                       & Ours              & \textbf{34.65} & \textbf{0.949} & \textbf{0.056}    \\ \hlineB{2.5}
\end{tabular}%
}
\label{tab:replica_psnr}
\end{table}
\footnotetext[2]{SplaTAM encounters memory issues in several scenes in Replica, so we only include the remaining results (office2/3/4 and room0) for reference.}

\begin{table}[]
\caption{Comparisons of Mapping Efficiency and Memory Cost}
\centering
\resizebox{0.9\columnwidth}{!}{%
\begin{tabular}{c|cccc}
\hlineB{2.5}
Dataset                    & Method           & \begin{tabular}[c]{@{}c@{}}Mapping\\ FPS\end{tabular} & \begin{tabular}[c]{@{}c@{}}Model size\\ (MB)\end{tabular} & \begin{tabular}[c]{@{}c@{}}GPU memory\\ usage (MB)\end{tabular} \\ \hlineB{2.5}
\multirow{3}{*}{ScanNet++} & SplaTAM          & 0.19                                                  & 206.3                                                     & 4417                                                            \\
                           & RTG-SLAM         & 1.29                                                  & 111.8                                                     & 3906                                                            \\
                           & Ours             & \textbf{6.14}                                         & \textbf{29.3}                                             & \textbf{2810}                                                   \\ \hlineB{2.5}
\multirow{3}{*}{Replica}   & SplaTAM$^\dagger$ & (0.14)                                                & (327.5)                                                   & (11132)                                                         \\
                           & RTG-SLAM         & 8.34                                                  & 62.6                                                      & \textbf{5105}                                                   \\
                           & Ours             & \textbf{9.73}                                         & \textbf{40.1}                                             & 7255                                                            \\ \hlineB{2.5}
\end{tabular}%
}
\label{tab:exp_efficiency}
\end{table}

\subsection{Ablation Study}
\label{sec:exp_ablation}
\subsubsection{Effect of TSDF Voxel Size}
In our setup, the size of the voxels is crucial not only for the TSDF fusion process but also for the Gaussian initialization stage. Each voxel can contain at most one Gaussian. Hence, using larger voxels results in fewer primitives, reducing computational demands and memory requirements, but leads to less photo-realistic reconstructions and less accurate TSDF grids as indicated in Table~\ref{tab:voxel_size}. Therefore, choosing the appropriate voxel size is finding a balance between map quality and computational efficiency.

\subsubsection{Effect of Quadtree Threshold}
As previously shown in Fig.~\ref{fig:qtree}, stricter quadtree thresholds can better preserve finer details. This is further confirmed in Table~\ref{tab:qtree_thresh}, where using a 10x smaller threshold achieves higher rendering quality. However, this comes at the cost of slower mapping speed and a larger model size due to the increased number of 3D Gaussians in the scene.

\subsubsection{Effect of Global Optimization}
We evaluate the effect of global optimization by changing the number of iterations applied after scanning and report the results in Table~\ref{tab:global_opt}. The more iterations of global optimization we use, the better rendering performance we can get. Notably, it only takes 1 minute to run 10 iterations of global optimization on a keyframe list with an average of 340 images, while the improvements in visual quality are significant. Therefore, we highly recommend performing several iterations of global optimization after scanning whenever time permits.

\begin{table}[]
\caption{Ablation Study of TSDF Voxel Size on ScanNet++}
\centering
\resizebox{\columnwidth}{!}{%
\begin{tabular}{cc|ccccc}
\hlineB{2.5}
Voxel size           & Data splits   & PSNR$\uparrow$ & SSIM$\uparrow$ & LPIPS$\downarrow$ & \begin{tabular}[c]{@{}c@{}}Mapping\\ FPS\end{tabular} & \begin{tabular}[c]{@{}c@{}}Model size\\ (MB)\end{tabular} \\ \hlineB{2.5}
\multirow{2}{*}{1cm} & Training view & \textbf{28.84} & \textbf{0.897} & \textbf{0.138}    & \multirow{2}{*}{6.14}                                 & \multirow{2}{*}{29.3}                                     \\
                     & Novel view    & \textbf{25.45} & \textbf{0.848} & \textbf{0.216}    &                                                       &                                                           \\ \hline
\multirow{2}{*}{5cm} & Training view & 28.71          & 0.894          & 0.143             & \multirow{2}{*}{\textbf{8.97}}                        & \multirow{2}{*}{\textbf{21.4}}                            \\
                     & Novel view    & 25.31          & 0.846          & 0.220             &                                                       &                                                           \\ \hlineB{2.5}
\end{tabular}
}
\label{tab:voxel_size}
\end{table}

\begin{table}[]
\caption{Ablation Study of Quadtree Threshold on ScanNet++}
\centering
\resizebox{\columnwidth}{!}{%
\begin{tabular}{cc|ccccc}
\hlineB{2.5}
\begin{tabular}[c]{@{}c@{}}Quadtree\\ threshold\end{tabular} & Data splits   & PSNR$\uparrow$ & SSIM$\uparrow$ & LPIPS$\downarrow$ & \begin{tabular}[c]{@{}c@{}}Mapping\\ FPS\end{tabular} & \begin{tabular}[c]{@{}c@{}}Model size\\ (MB)\end{tabular} \\ \hlineB{2.5}
\multirow{2}{*}{0.01}                                        & Training view & \textbf{29.23} & \textbf{0.901} & \textbf{0.126}    & \multirow{2}{*}{5.09}                                 & \multirow{2}{*}{48.8}                                     \\
                                                             & Novel view    & \textbf{25.96} & \textbf{0.853} & \textbf{0.204}    &                                                       &                                                           \\ \hline
\multirow{2}{*}{0.1}                                         & Training view & 28.84          & 0.897          & 0.138             & \multirow{2}{*}{\textbf{6.14}}                        & \multirow{2}{*}{\textbf{29.3}}                            \\
                                                             & Novel view    & 25.45          & 0.848          & 0.216             &                                                       &                                                           \\ \hlineB{2.5}
\end{tabular}
}
\label{tab:qtree_thresh}
\end{table}

\begin{table}[]
\caption{Ablation Study of Global Optimization on ScanNet++}
\centering
\resizebox{\columnwidth}{!}{%
\begin{tabular}{cc|cccc}
\hlineB{2.5}
\begin{tabular}[c]{@{}c@{}}\#Global opt.\\ iterations\end{tabular} & Data splits   & PSNR$\uparrow$ & SSIM$\uparrow$ & LPIPS$\downarrow$ & \begin{tabular}[c]{@{}c@{}}Global opt.\\ time (s)\end{tabular} \\ \hlineB{2.5}
\multirow{2}{*}{0}                                                 & Training view & 24.99          & 0.839          & 0.243             & \multirow{2}{*}{\textbf{0.0}}                                  \\
                                                                   & Novel view    & 22.76          & 0.794          & 0.313             &                                                                \\ \hline
\multirow{2}{*}{10}                                                & Training view & 28.84          & 0.897          & 0.138             & \multirow{2}{*}{60.1}                                          \\
                                                                   & Novel view    & 25.45          & 0.848          & 0.216             &                                                                \\ \hline
\multirow{2}{*}{20}                                                & Training view & \textbf{29.50} & \textbf{0.906} & \textbf{0.117}    & \multirow{2}{*}{120.7}                                         \\
                                                                   & Novel view    & \textbf{25.87} & \textbf{0.856} & \textbf{0.194}    &                                                                \\ \hlineB{2.5}
\end{tabular}
}
\label{tab:global_opt}
\end{table}

\begin{table}[]
\caption{Ablation Study of Random Keyframe Optimization on Replica}
\centering
\resizebox{\columnwidth}{!}{%
\begin{tabular}{c|ccccc}
\hlineB{2.5}
Optimization strategy    & PSNR$\uparrow$ & SSIM$\uparrow$ & LPIPS$\downarrow$ & \begin{tabular}[c]{@{}c@{}}Mapping\\ FPS\end{tabular} & \begin{tabular}[c]{@{}c@{}}GPU memory\\ usage (MB)\end{tabular} \\ \hlineB{2.5}
w/o random keyframe opt. & 24.22          & 0.849          & 0.150             & 9.70                                                  & 7100                                                            \\ \hline
w random keyframe opt.   & \textbf{28.64} & \textbf{0.900} & \textbf{0.103}    & 9.74                                                  & 7092                                                            \\ \hlineB{2.5}
\end{tabular}
}
\label{tab:random_kf_opt}
\end{table}

\subsubsection{Effect of Random Keyframe Optimization}
We conduct ablation studies on our keyframe maintenance strategy using Replica dataset, as it contains more frames in each scene compared to ScanNet++ dataset (2000 vs. $\sim$340). We perform 5 iterations of optimization for each frame, in contrast to the standard setting of 5 iterations for keyframes and 3 for non-keyframes with an additional 2 iterations for random keyframe optimization. Note that we don't employ any global optimization here to fully manifest the effect of random keyframe optimization. 

From Table~\ref{tab:random_kf_opt}, we can see that the usage of random keyframe optimization enhances visual quality by a large margin without impacting mapping speed or memory cost. This improvement is due to the fact that revisiting keyframes in the maintained list can help prevent the system from forgetting previously scanned scenes. We perform this process only when the current frame is not a keyframe so that we can reduce the number of iterations needed for the current frame, allowing resources to be allocated to random keyframe optimization without increasing computational time or memory usage.

\subsubsection{Effect of Keyframe Threshold}
We test different keyframe thresholds as shown in Table~\ref{tab:kf_thresh}. When loosening the keyframe threshold, the keyframe list will also expand. Consequently, it will generate better rendering results at the expense of increased GPU memory usage and longer global optimization time.

\begin{table}[]
\caption{Ablation Study of Keyframe Threshold on Replica}
\centering
\resizebox{\columnwidth}{!}{%
\begin{tabular}{cc|ccccc}
\hlineB{2.5}
\begin{tabular}[c]{@{}c@{}}Keyframe\\ threshold\end{tabular} & \begin{tabular}[c]{@{}c@{}}\#Avg.\\ keyframes\end{tabular} & PSNR$\uparrow$ & SSIM$\uparrow$ & LPIPS$\downarrow$ & \begin{tabular}[c]{@{}c@{}}GPU memory\\ usage (MB)\end{tabular} & \begin{tabular}[c]{@{}c@{}}Global opt.\\ time (s)\end{tabular} \\ \hlineB{2.5}
40                                                           & 655                                                        & \textbf{35.44} & \textbf{0.955} & \textbf{0.051}    & 7915                                                            & 167.3                                                          \\ \hline
50                                                           & 594                                                        & 34.65          & 0.949          & 0.056             & 7255                                                            & 151.5                                                          \\ \hline
90                                                           & 414                                                        & 33.26          & 0.941          & 0.065             & \textbf{5163}                                                   & \textbf{105.5}                                                 \\ \hlineB{2.5}
\end{tabular}
}
\label{tab:kf_thresh}
\end{table}

\section{Conclusion}
\label{sec:conclusion}
In this paper, we have introduced an online RGB-D mapping system that takes advantage of volumetric grids and Gaussian splatting to simultaneously create a geometrically accurate TSDF map and a photo-realistic Gaussian map. Our Gaussian map remains compact as we use a quadtree data structure to recursively subdivide each input RGB image into cells of varying sizes based on contrast, where the number of cells is much smaller than the total number of pixels. Before initializing a new Gaussian, we check the neighborhood of the back-projected cell center to prevent unnecessary map expansion. This approach effectively limits the number of Gaussian parameters that need updating during optimization. With a tailored keyframe maintenance strategy, we can optimize the 3D Gaussian map in real time. Experiments on both synthetic and real-world datasets demonstrate that our method, GSFusion, strikes a balance between visual quality and computational efficiency, outperforming previous Gaussian SLAM methods. We have also conducted comprehensive ablation studies to explore how different design choices impact the system's performance, providing insights for maximizing its use in various applications. In the future, we plan to explore the integration of our system with multi-resolution volumetric grids for large-scale scenarios.

\section*{Acknowledgment}

This work was supported by the EU project AUTOASSESS. The authors would like to thank Simon Boche and Sebastián Barbas Laina for their assistance in collecting and processing drone data. We also extend our gratitude to Sotiris Papatheodorou for his valuable discussions and support with the Supereight2 software.

\ifCLASSOPTIONcaptionsoff
  \newpage
\fi

\bibliographystyle{IEEEtran}
\bibliography{root}

\end{document}